\documentclass[APA,LATO1COL]{WileyNJD-v2}

\articletype{Article Type}%

\received{26 April 2016}
\revised{6 June 2016}
\accepted{6 June 2016}

\usepackage{subcaption}
\newcommand*\mean[1]{\overline{#1}}
\newcommand{\modelName}{Nested Variational Autoencoder}
\newcommand{\modelNameLower}{nested variational autoencoder}
\newcommand{\modelAbbr}{N-VAE}
\newcommand{\instituteName}{Ho Chi Minh City University of Technology}
\begin{document}

\title{\modelName{} for Topic Modeling on Microtexts with Word Vectors}

\author[1]{Trung Trinh}
\author[1]{Tho Quan}
\author[1]{Trung Mai}

\authormark{Trinh \textsc{et al}}

\address[1]{\orgdiv{Faculty of Computer Science and Engineering}, \orgname{\instituteName}, \orgaddress{\state{Ho Chi Minh}, \country{Vietnam}}}

\corres{Tho Quan, \orgdiv{Faculty of Computer Science and Engineering}, \orgname{\instituteName}, \orgaddress{\state{Ho Chi Minh}, \country{Vietnam}}.\\ \email{qttho@hcmut.edu.vn}}


\abstract[Summary]{Most of the information on the Internet is represented in the form of \textit{microtexts}, which are short text snippets such as news headlines or tweets.
These sources of information are abundant, and mining these data could uncover meaningful insights.
Topic modeling is one of the popular methods to extract knowledge from a collection of documents; however, conventional topic models such as latent Dirichlet allocation (LDA) are unable to perform well on short documents, mostly due to the scarcity of word co-occurrence statistics embedded in the data. 
The objective of our research is to create a topic model that can achieve great performances on microtexts while requiring a small runtime for scalability to large datasets.
To solve the lack of information of microtexts, we allow our method to take advantage of word embeddings for additional knowledge of relationships between words. 
For speed and scalability, we apply autoencoding variational Bayes, an algorithm that can perform efficient black-box inference in probabilistic models.
The result of our work is a novel topic model called the \modelNameLower{}, which is a distribution that takes into account word vectors and is parameterized by a neural network architecture. 
For optimization, the model is trained to approximate the posterior distribution of the original LDA model.
Experiments show the improvements of our model on microtexts as well as its runtime advantage.}



\maketitle

\section{Introduction}\label{sec:intro}
The ubiquity of microtexts, which is due to the emergence of social media sites such as Facebook and Twitter, has become an increasingly valuable asset for mining information about the real world. In health care, by monitoring information posted by users on social networks, one can observe the status of public health \citep{paul_social_2015}. nEmesis \citep{sadilek_deploying_2017} is a system deployed in combination with social media to prevent foodborne illness. On a broader scale, social platform data can be used to observe public ideas \citep{kennedy_social_2015} and offer emergency service \citep{pandey_2018}. The RSC system \citep{costa_2015} is a system that monitors temporal human activities. More sophisticated mining tasks can be employed to detect fake news \citep{Shu:2017:FND:3137597.3137600} or to detect themes on social media \citep{lazard2015}.

Many of these applications could be distilled to inferring topics from these sources of information. Probabilistic models such as probabilistic latent semantic analysis (PLSA) \citep{hofmann_probabilistic_1999} and latent Dirichlet allocation (LDA) \citep{blei_latent_2003} have been successfully applied to long texts. These models operate on the assumption that each document comprises a small number of topics, each of which in turn consists of a subset of words. The topic proportions of each document as well as the distribution over the vocabulary of each topic are then learned from the corpus using statistical methods such as Gibbs sampling \citep{griffiths_finding_2004} or variational inference \citep{blei_latent_2003}. The effectiveness of these models strongly depends on the patterns of word cooccurrences within the corpus, which is fully and correctly represented in a large corpus of long texts. The sparsity of microtexts, on the other hand, has proven to be inadequate in presenting the relationship between words. Thus, microtexts remain challenging for these conventional topic modeling methods. One way to alleviate this problem is to introduce additional information that could help the model to uncover the true semantic relationships between words. Such methods include utilizing search results for matching similar text snippets \citep{sahami_web-based_2006}, using a  knowledgebase for microtext conceptualization \citep{song_short_2011}, and leveraging auxiliary long texts to enhance microtext clustering performance \citep{jin_transferring_2011}.

Recently, the introduction of word embeddings \citep{mikolov_efficient_2013,NIPS2013_5021,pennington_glove:_2014} has led to improvements in many natural language processing (NLP) tasks due to their ability to capture semantic relationships between words in a distributed fashion. In this model, each word is represented using a dense vector that contains the semantic and syntactic information about that word, and similar words tend to stay close to each other in the embedding space. One could expect that these latent features of words could be used to improve the performance of topic modeling on microtexts. In fact, there have been some studies that incorporate word vectors with available topic modeling methods such as LDA and have shown promising results \citep{das_gaussian_2015,nguyen_improving_2015}. While the improvements are remarkable, these papers use Gibb sampling as the method to infer the parameters of the model, and although it has been proven theoretically to produce the best results compared to other methods, it requires a long time to converge. 

With the advancement of deep learning in recent years, these methods have been used to replace the traditional mean field approximator in variational inference, which has led to the invention of variational autoencoders (VAEs) \citep{kingma_auto-encoding_2013,rezende_stochastic_2014}. Most studies on VAEs employ the Gaussian distribution since the reparameterization trick (RT) for the Gaussian distribution is readily available \citep{kingma_auto-encoding_2013,rezende_stochastic_2014}. To approximate the Dirichlet prior of LDA, \cite{srivastava_autoencoding_2017} used a logistic normal distribution to approximate the Laplace approximation of the Dirichlet prior on the softmax basis. Their study provides evidence that VAEs could be effective for topic modeling, showing better results than traditional LDA with Gibbs sampling while enjoying much faster convergence.

The recent development in RT allows more distributions to be used in VAEs \citep{figurnov_implicit_2018,jang_categorical_2016,maddison_concrete_2016}. This has opened up new possibilities for VAEs to better approximate complex probabilistic models, which include LDA. The experiment in \cite{figurnov_implicit_2018} indicates that using the Dirichlet prior directly in VAE produces lower perplexity than that of \cite{srivastava_autoencoding_2017}, and the model also has the ability to further fine-tune the parameters of the Dirichlet prior, which has proven to greatly improve the performance of LDA \citep{wallach_rethinking_2009}.

In this paper, we introduce a new VAE model that can leverage additional information from word embeddings to perform topic modeling on microtexts. Our motivation originates from the need of an algorithm that is effective in detecting latent topics in a corpus of short texts, which is achieved through the assistance of word embeddings, and has a good performance in order to scale up to a large dataset by using amortized variational inference with a neural network. However, this approach poses two major challenges:
\begin{itemize}
\item One must design a model that can appropriately use word vectors to their advantage.
\item The original VAE approach only supports approximating the Gaussian distribution. Meanwhile, the topics generated for each document by LDA follow the Dirichlet distribution, and the relationships between each topic and each word are categorical.
\end{itemize}
To overcome these difficulties, we first posit a probability distribution, denoted \emph{\textbf{q}}, that factors into two conditional distributions: the first one is the topic distribution of each word conditioned on its embedding and current context; the second one is the topic distribution of each document based on all the topic assignments of all the words in that document. We then approximate this probability distribution using a neural network with a purposely designed architecture.
The resulting network is a 2-layered nested structure of latent variables, which we aptly name \emph{\modelNameLower{}} (\emph{\modelAbbr{}}). To find the parameter of the neural network, we minimize the Kullback-Leibler divergence between \emph{\textbf{q}} and the posterior probability distribution of the LDA model by maximizing the corresponding evidence lower bound (ELBO) function, as normally done in the variational inference method. We use the Gumbel-Softmax trick \citep{jang_categorical_2016,maddison_concrete_2016} and the technique described by \cite{figurnov_implicit_2018}, respectively, to optimize the parameters of the the word-to-topic distribution (a categorical distribution) and the document-to-topic distribution (a Dirichlet distribution) in \emph{\textbf{q}} using stochastic gradient descent.

Compared to other LDA-based approaches for topic modeling, our approach enjoys the following advantages.

\begin{itemize}
\item Since word vectors are encoded with rich semantic information, we can potentially generate more meaningful topics. Moreover, the word vectors can be further fine-tuned in the training process through backpropagation.
\item Using a neural network to parameterize the approximator in the variational inference setting gives us a shorter convergence time (since neural networks allow concurent processing) and smaller memory requirements (by using mini-batch gradient descent for training).
\end{itemize}

Indeed, experiments on various datasets of microtexts show that our model produces much better results on document clustering tasks and competitive results on topic coherence evaluation compared to other methods, while requiring a much smaller amount of time for convergence. Our implementation is available at
\url{https://github.com/trungtrinh44/N-VAE}.

\section{Related Works}\label{sec:related_words}
With the ability to infer latent topics from a collection of documents, topic modeling has many interesting applications in various fields: detecting themes in historical newspapers \citep{Newman:2006:PTD:1124169.1124187,yang_2011_historical}, studying scholarly literature \citep{Mimno:2012:CHD:2160165.2160168,Goldstone:2014,Riddell:2014}, and analyzing biological data \citep{liu_overview_2016}.
In terms of social media, the topic model has been utilized for identifying influential users on Twitter \citep{Weng:2010:TFT:1718487.1718520}, detecting communities \citep{tamimi_community_2017}, and comparing topics of interest between different regions \citep{Yin:2011:GTD:1963405.1963443}.
Even though popular methods such as LDA \citep{blei_latent_2003} have been successfully applied to collections of long texts, microtexts remain a huge challenge for these models.
The sparsity of microtexts provides insufficient information on word cooccurrences required to infer the topic-to-word and document-to-topic distributions.
Therefore, various solutions have been proposed to overcome this problem. 

One way to solve the lack of signals from microtexts is to aggregate them into a large document using some heuristics.
This method is more commonly used for social media data, by leveraging the structure of the network and the relationship between the data.
\cite{Hong:2010:EST:1964858.1964870} explore the author-based aggregate scheme, i.e., combining tweets of each user to form a document.
Such aggregation is also used in \cite{Weng:2010:TFT:1718487.1718520}.
\cite{Mehrotra:2013:ILT:2484028.2484166}, further introducing other aggregate schemes such as pooling tweets containing similar hashtags (Hashtag-based pooling), pooling tweets posted within the same hour to detect major events (temporal pooling), etc.
These methods have been shown to improve the performance of LDA on tweet data, but they depend on the specific properties of the dataset, which means they cannot be generalized to other use cases of microtext-based topic modeling. 

Another direction is to invent new topic models that are more suitable for the characteristics of microtexts.
\cite{Yin:2014:DMM:2623330.2623715} study the effectiveness of the Dirichlet multinomial mixture model on microtext clustering.
\cite{Yan_biterm:2013} introduce the biterm topic model whose generative process is specifically designed to adapt to the sparseness of microtexts.
\cite{Quan:2015:SST:2832415.2832564} propose a model with a two-phase generative process, where the first phase is similar to the LDA model while the second phase assumes that each text snippet is derived from a hidden pseudo-document.
While these models are more effective than LDA on microtexts, they can only use the limited information provided by the training corpus, which may not truly reflect the semantic relationships between words within the vocabulary.

To improve the results of topic models on microtexts, auxiliary information could be provided to augment or act as an alternative to the word cooccurrence statistics from the current corpus.
\cite{petterson_word_2010} use additional information on word similarities from thesauri and dictionaries to help place synonyms into the same topic.
\cite{song_short_2011} employ a probabilistic knowledgebase to improve short text comprehension.
\cite{jin_transferring_2011} develop an extension to the LDA model called dual LDA, where the target topics of a short text dataset are jointly learned with the supporting topics from an external collection of long texts.
Other simpler approaches include training a topic model on a very large and universal corpus and then using the trained model to infer topics on a microtext corpus \citep{phan_hidden_2011}. 

Word embedding \citep{mikolov_efficient_2013,NIPS2013_5021,pennington_glove:_2014} is a distributed representation of words and encodes the semantic relationships between them, which could be useful in aiding topic modeling.
Some studies on using word embeddings in topic modeling have resulted in better performances on microtexts.
\cite{qiang_topic_2017} combine word embeddings with a text aggregation method and the Markov random field regularized model.
\cite{das_gaussian_2015,nguyen_improving_2015} integrate LDA with word vectors via adding new components to the original model.
However, these papers only explore Gibbs sampling as the training method, which is slow and can not scale up to a large dataset, rendering them impractical for real use cases.

Variational inference with the mean field approximator could be used to train the LDA model \citep{blei_latent_2003,Teh:2006:CVB:2976456.2976626}, which is faster than Gibbs sampling.
\cite{kingma_auto-encoding_2013,rezende_stochastic_2014} replace the mean field approximator with the neural network, resulting in the VAE.
\cite{srivastava_autoencoding_2017} present a VAE model for topic modeling based on LDA that has a much faster training and inference time while maintaining a competitive performance to the original LDA model trained using conventional methods.
However, this model suffers from the component collapsing problem when training on microtexts.

We take inspiration from those previous works and create a new model that employs a neural network architecture to approximate the posterior distribution of LDA and can benefit from word embeddings for knowledge of semantic relationships between words.
We believe that using auxiliary information for assistance is the most general solution to tackle the problems of constructing a topic model on microtexts.
This method not only works better on microtexts but could also enjoy improvements on long texts.
We choose the stochastic variational inference with a neural network approximator because we want our model to scale up to large datasets without needing a long time to converge and consuming a large amount of memory.

\section{Background}\label{sec:background}
\subsection{Word Embeddings}
Word embeddings encode the information of each word using a dense vector.
Two popular unsupervised methods for training word embeddings are Word2Vec \citep{mikolov_efficient_2013,NIPS2013_5021} and GloVe \citep{pennington_glove:_2014}.
Word2Vec uses a shallow neural network consisting of an input layer, a projection layer and an output layer to predict neighboring words.
There are two versions of Word2Vec: continuous bag-of-words (CBOW) and skip-gram.
The CBOW model attempts to predict a word based on its context, i.e., its surrounding words, while the skip-gram model does the inverse, predicting context words based on the current word input.

Instead of learning the relationships between one word and its neighbors similar to Word2Vec, the GloVe method trains a log-bilinear regression model directly on the matrix of the global word cooccurrence statistics of a corpus.
Its objective is to approximate the log probability of the cooccurrence of two words using the dot product of their word vectors.

Both Word2Vec and GloVe produce vectors that can encapsulate semantic relationships between words, which are usually encoded in the differences between two vectors. 
For example, $\text{vec(Vietnam)} - \text{vec(Hanoi)} \approx \text{vec(France)} - \text{vec(Paris)} \approx  \text{vec(Germany)} - \text{vec(Berlin)}$.
Likewise, words that are near each other in the embedding space usually have certain similarities in their meanings.

\subsection{Latent Dirichlet Allocation}
\begin{algorithm}
 \caption{The generative process of LDA}
 \label{algo:lda}
 \begin{algorithmic}
 \For {each document $d$}
    \State Sample a document-to-topic distribution $\theta_d \sim \text{Dirichlet}(\alpha)$\;
    \For {each word $i$ in document $d$}
        \State Sample a topic $z_{d,i} \sim \text{Categorical}(\theta_d)$\;
        \State Sample a word $w_{d,i} \sim \text{Categorical}(\beta_{z_{d,i}})$\;
    \EndFor
 \EndFor
 \end{algorithmic}
\end{algorithm}
\begin{figure}
  \centering
  \includegraphics[width=.7\textwidth]{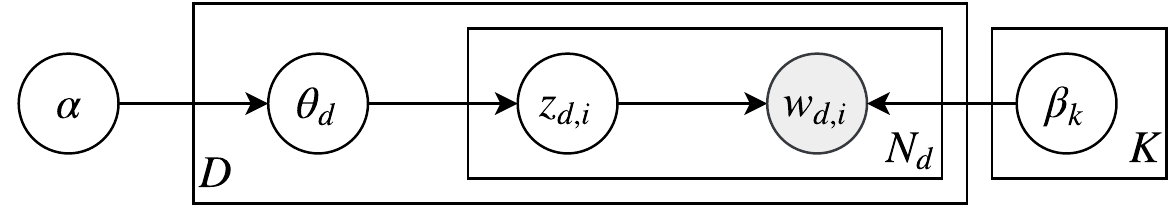}
  \caption{The LDA graphical model.}
  \label{fig:lda}
\end{figure}
LDA \citep{blei_latent_2003} is a probabilistic generative model popularly used to extract latent topics from a collection of documents. The model assumes that each document is a mixture of topics and each topic is a probability distribution over a fixed vocabulary. The generative process of LDA is described in Algorithm \ref{algo:lda} and depicted in Figure \ref{fig:lda}, where:
\begin{description}
    \item[$\bullet$] $\beta_k$ represents the parameters of the categorical distribution over the vocabulary of topic $k$ and $K$ is the number of topics; 
    \item[$\bullet$] $\theta_{d}$ represents the parameters of the categorical distribution over topics of document $d$ and $D$ is the number of documents; 
    \item[$\bullet$] $z_{d,i}$ is the topic assignment for the $i$-th word of document $d$; 
    \item[$\bullet$] $w_{d,i}$ is the $i$-th word of document $d$ and $N_d$ is the number of words in document $d$.
    \item[$\bullet$] $\alpha$ is the parameter of the Dirichlet prior distribution.
\end{description}

Subsequently, the joint probability distribution of LDA is defined as:
\begin{equation}
\begin{split}
    p(w,z,\theta|\alpha,\beta) = \prod_{d} p(\theta_d|\alpha) \prod_{i} p(z_{d,i}|\theta_d) p(w_{d,i}|\beta_{z_{d,i}})
\end{split}
\end{equation}

\subsection{Variational Inference}
Probabilistic generative models such as LDA require Bayesian inference methods to induce the values of their latent variables from the corresponding posterior distribution. Recall that in Bayes' theorem, $p(z|x)$ can be evaluated as
\begin{equation}
\label{eq:bayes}
\begin{split}
  p(z|x) = \frac{p(x|z)p(z)}{p(x)}.
\end{split}
\end{equation}

However, such a distribution is usually intractable, leading to the employment of approximation methods. One such method is variational inference, where a tractable distribution $q$ is used to approximate the true distribution $p$. To find $q$ that most resembles $p$, we find $q$ that minimizes the Kullback-Leibler (KL) divergence from $q$ to $p$. Given a model with evidence $x$ and a set of latent variables $z$, the KL divergence from $q(z)$ to the posterior distribution $p(z|x)$ is defined as:
\begin{equation}
\label{eq:kl_vi}
\begin{split}
    KL(q(z)||p(z|x)) &= -\mathbb{E}_{q(z)}[\log \frac{p(z|x)}{q(z)}] \\
    &= \mathbb{E}_{q(z)}[\log q(z) - \log p(z|x)]
\end{split}
\end{equation}

Since directly minimizing this function requires knowing how to calculate $p(z|x)$, which is intractable in the first place, we instead maximize the ELBO function:
\begin{equation}
    \begin{split}
        \text{ELBO} &= \mathbb{E}_{q(z)}[\log p(z,x) - \log q(z)] \\
        &= \mathbb{E}_{q(z)}[\log p(x) + \log p(z|x) - \log q(z)] \\
        &= \log p(x) - \mathbb{E}_{q(z)}[\log q(z) - \log p(z|x)] \\
        &= \log p(x) - KL(q(z)||p(z|x))
    \end{split}
\end{equation}

The name ELBO stems from the fact that the term is always smaller or equal to the log probability of the evidence, i.e., $\log p(x) \geq \text{ELBO}$ since $KL(q(z)||p(z|x))$ $\geq 0$. Because $\log p(x)$ is a constant with regard to the observed data $x$ and the parameters of $q$, maximizing the ELBO corresponds to minimizing the $KL(q(z)||p(z|x))$.

The chosen approximate posterior $q$ usually comes from the mean field family, where each latent variable is assumed to come from a distribution with its own parameters. The optimization process requires deriving the updating rules by hand analytically for the coordinate descent algorithm. The analytical solution only exists if the model is conjugate. LDA is one such model because of the conjugacy between the Dirichlet and the multinomial distributions. Hence, the largest limitation of mean field variational inference is that it can only be used with conjugate models.

\subsection{Variational Autoencoder and Reparameterization Tricks}
\begin{figure}
  \centering
  \includegraphics[width=.7\textwidth]{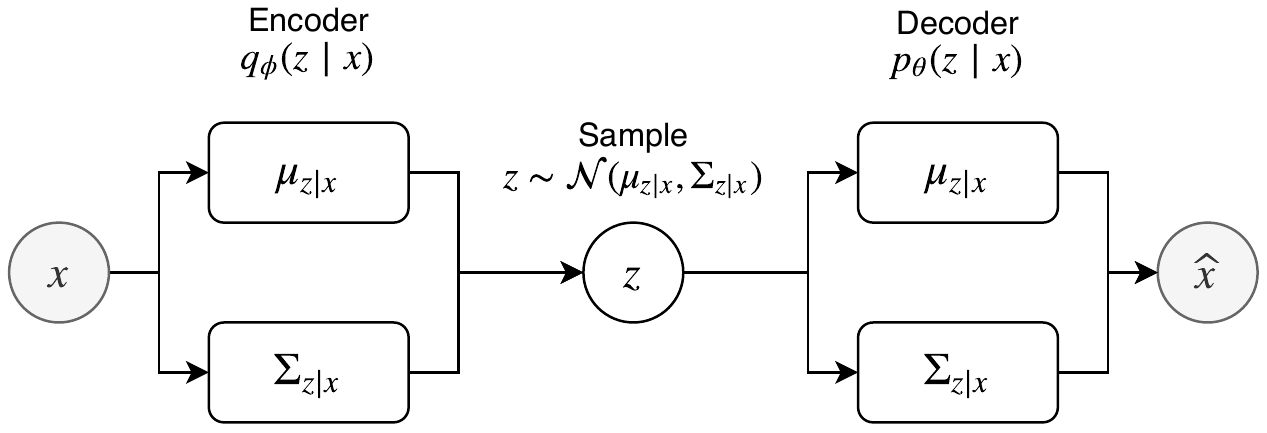}
  \caption{An example VAE with a Gaussian posterior and a Gaussian decoder.}
  \label{fig:vae_gaussian}
\end{figure}
One way to mitigate the limitation of mean field variational inference is to use a neural network to parameterize the approximate posterior $q$. This idea is explored in \cite{kingma_auto-encoding_2013,rezende_stochastic_2014}, which gives rise to a new class of models called the VAE. Essentially, the VAE consists of two parts: the \emph{encoder} $q_{\phi}(z|x)$ and the \emph{decoder} $p_{\theta}(x|z)$, where $\phi$ and $\theta$ denote the parameters of the encoder and decoder respectively. The encoder role is to map each of the input $x$ to its corresponding latent variables $z$, and the decoder role is to reconstruct $x$ from $z$. Ideally, we want the encoder $q_{\phi}(z|x)$ to act as if it is the true posterior $p_{\theta}(z|x)$ as much as possible. That goal is achieved by minimizing the KL-divergence between the two distributions, similar to the variational inference method. Under this setting, Equation \eqref{eq:kl_vi} is rewritten as:
\begin{equation}
    \label{eq:kl_vae}
    \begin{split}
    KL(q_{\phi}(z|x)||p_{\theta}(z|x))=\mathbb{E}_{q_{\phi}(z|x)}[\log q_{\phi}(z|x) - \log p_{\theta}(z|x)]
    \end{split}
\end{equation}
and the new ELBO is:
\begin{equation}
    \label{eq:elbo_vae}
    \begin{split}
        \text{ELBO} &= \mathbb{E}_{q_{\phi}(z|x)}[\log p_{\theta}(z,x) - \log q_{\phi}(z|x)] \\
        &= -KL(q_{\phi}(z|x)||p_{\theta}(z)) + \mathbb{E}_{q_{\phi}(z|x)}[\log p_{\theta}(x|z)]
    \end{split}
\end{equation}

Figure \ref{fig:vae_gaussian} depicts a VAE whose latent variables $z$ are assumed to be generated from a Gaussian distribution.

The KL term in Equation \eqref{eq:elbo_vae} usually has a closed-form expression. However, samplings need to be done in order to approximate the expectation term. There is a direct formula to calculate the gradients of the expectation with respect to the variational parameters, but this method yields gradients with very high variance, making training impossible. Therefore, both papers introduce an alternative called the RT. This method uses a differentiable and invertible function $g_\phi$ such that $\varepsilon = g_{\phi}(z)$ and $z = g_{\phi}^{-1}(\varepsilon)$. Here, $\varepsilon \sim p(\varepsilon)$ is called a noise variable, and this transformation helps remove the dependence of the sampling process on the variational parameters. Instead of sampling directly from $q_\phi(z|x)$, we now draw $\varepsilon$ from its distribution and calculate the corresponding $z$. If $q_\phi(z|x)$ happens to be a Gaussian distribution $\mathcal{N}(\mu,\sigma)$, we can choose $p(\varepsilon) = \mathcal{N}(0,1)$ and $g_{\phi}(z)=(z-\mu)/\sigma$, i.e., $z=\mu + \sigma \varepsilon$. Unfortunately, finding a pair of $p(\varepsilon)$ and a function $g_\phi$ is not a trivial task for other distributions, such as the Dirichlet or categorical distributions used in LDA. One universal solution for every continuous distribution is to consider $p(\varepsilon) = \text{Uniform}(0,1)$ and $g_{\phi}(z)=F_\phi(z|x)$ where $F_\phi(z|x)$ is the cumulative distribution function of $q_\phi(z|x)$; however, to calculate $z$, we need to find the inverse of $F_\phi(z|x)$, which could be a very complicated process. The work in \cite{figurnov_implicit_2018} proposes a novel way to alleviate this problem called implicit reparameterization, which in contrast to the method discussed so far does not need to find the inverse of $g_\phi$. This solution enables the RT to be used with a variety of continuous distributions, including the Dirichlet distribution. For discrete cases such as the categorical distribution, we use the Gumbel-Softmax \citep{jang_categorical_2016,maddison_concrete_2016} as a continuous approximation, which also permits backpropagation to update the distribution parameters.

\section{Nested Variational Autoencoder (\modelAbbr{}) for Topic Modeling on Microtext with Word Vectors}
In this section, we introduce \emph{\modelAbbr{}}, a novel VAE architecture for topic modeling with word embeddings. Our approach differs from other works on combining word embeddings and LDA since instead of replacing or extending components of the original LDA model with elements that are compatible with word vectors \citep{das_gaussian_2015,nguyen_improving_2015}, we propose a variational approximation \emph{\textbf{q}} to the posterior of the LDA model that considers word embeddings as one of its parameters. In the following subsections, we will introduce the formulation of \emph{\textbf{q}}, the derivation of the ELBO as the objective function and the translation of \emph{\textbf{q}} to a neural network architecture. We will denote $V$, $K$, $D$ as the vocabulary size, the number of topics and the word embedding size, respectively.

\subsection{The proposed distribution \emph{q}}\label{distribution_q}
\begin{figure}
  \centering
  \includegraphics[width=.7\textwidth]{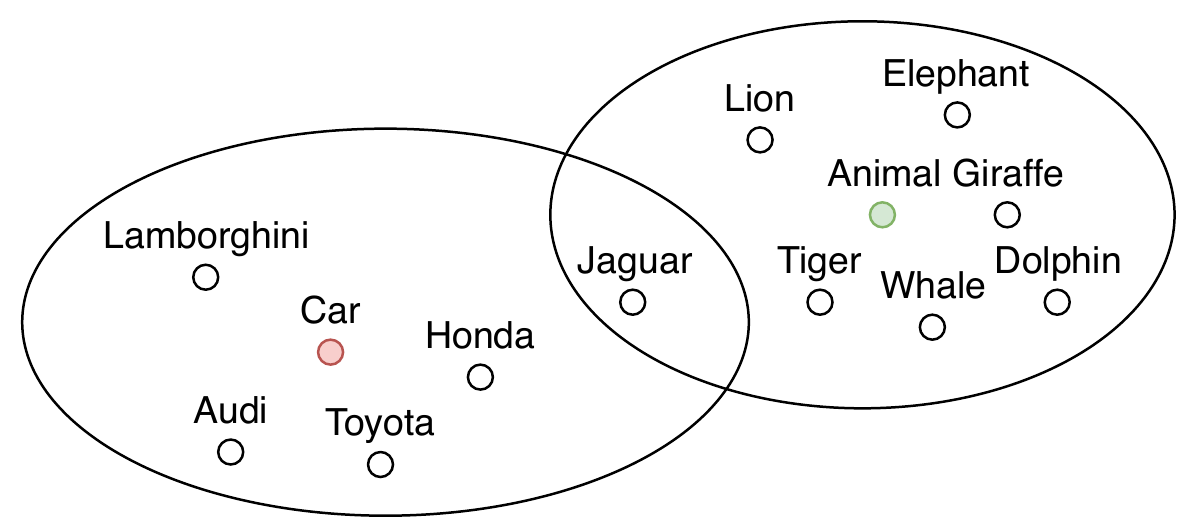}
  \caption{An example of word embeddings with two topic vectors: \emph{Car} and \emph{Animal}. Here, the word \emph{Jaguar} is a homonym which could mean an animal or a name of a car manufacturer.}
  \label{fig:wv_ex}
\end{figure}
To design the variational distribution \emph{\textbf{q}}, we make two assumptions:
\begin{itemize}
    \item A coherent topic should be a group of words whose vectors are close to each other in the embedding space.
    \item From the embedding and the context of a word, we could decide which topic that the word belongs to.
\end{itemize}
To realize both assumptions, we parameterize the word-to-topic distribution - a categorical distribution - using word embeddings and a set of vector representations of the topic (one vector for each topic). More specifically, for word $i$, topic $t$ and document $d$, the log probability of the unnormalized word-to-topic distribution is a sum of two factors: the first one, denoted $s_{d,i}^t$, is the dot product between the word embedding $\omega_i$ and the topic embedding $\rho_t$; the second one, denoted $c_d^t$, is a function $g_t$ that receives the context representation of that word in document $d$ and outputs the coefficient of topic $t$:
\begin{equation}
\label{eq:word2topic}
\begin{split}
    s_{d,i}^t &= \omega_{i}^d \cdot \rho_t \\
    c_d^t &= g_t(\mean{\omega^{d}}) \\
    \log \Tilde{q}(z_{d,i} = t | \rho_t, \omega^d) &= \pi_{d,i}^t = s_{d,i}^t + c_d^t
    \\
    q(z_{d,i} = t | \rho_t, \omega^d) &= \frac{\exp(\pi_{d,i}^t)}{\sum_t \exp(\pi_{d,i}^t)}
\end{split}
\end{equation}
where $z_{d,i}$ is the topic assigned to word $i$ in document $d$, and $\omega^d$ is the matrix of all the word embeddings within document $d$, $\mean{\omega^{d}}$ is the mean of all the word vectors in document $d$. Here, we represent the context of a word in a document as a the mean of word vectors in that document since we want the representation to be agnostic to the document's length. This strategy works well in practice.

The intuition behind this formulation is as follows:
\begin{itemize}
    \item For $s_{d,i}^t$, since a topic should be a group of neighboring word embeddings, the dot product between each word vector in the topic and the topic vector should be large. Therefore, this term will put a high probability on a topic whose vector is near the vector of the current word. This property is illustrated in Figure \ref{fig:wv_ex}.
    \item To account for homonyms, we introduce the term $c_d^t$, since these words have different meanings in different contexts and thus will belong to different topics. For example, the word \emph{Jaguar} in Figure \ref{fig:wv_ex} could indicate an animal or a car manufacturer depending on the current context.
\end{itemize}

In practice, we use the Gumbel-Softmax estimator to approximate the categorical distribution. Therefore, Equation \eqref{eq:word2topic} is rewritten as:
\begin{equation}\label{word2topic_gumbel}
\begin{split}
    q(z_{d,i} = t | \rho_t, \omega^d) = \frac{\exp(\pi_{d,i}^t/\tau)}{\sum_t \exp(\pi_{d,i}^t/\tau)} = \mu_{d,i}^t
\end{split}
\end{equation}
where $\tau$ is called the temperature. This term controls how much the continuous approximation resembles the real categorical distribution, with $\tau$ approaching $0$ resulting in a more discrete-like sample. When training, we anneal the temperature from $1$ to a predefined minimum value.

With each word in a document now assigned a topic based on the distribution defined in Equation \eqref{eq:word2topic}, we calculate the topic proportion of document $d$ as $q(\theta_d|z_d) = Dirichlet(\nu_d)$, where $\nu_d$ is defined as:
\begin{equation}
\begin{split}
        \nu_d = \texttt{softplus}(\eta_d * a + b)
\end{split}
\end{equation}
where 
\begin{equation}\label{eta}
\begin{split}
    \eta_d=[\eta_{d,0},\eta_{d,1},...,\eta_{d,K-1}]
\end{split}
\end{equation}
is a $K$-dimension vector and $\eta_{d,i}$ is the number of words in document $d$ assigned the $i$-th topic, $a$ and $b$ are scalars that map $\eta_d$ to the appropriate values for the parameters of the Dirichlet distribution, and $\texttt{softplus}=\log (1+\exp(x))$. The use of $\texttt{softplus}$ is to make sure that $\nu_d$ is always positive, a requirement for Dirichlet parameters.

Overall, the variational posterior \emph{\textbf{q}} is defined as:
\begin{equation}
\label{q_definition}
\begin{split}
    q(\theta, z|\rho,\omega)&=\prod_d q(\theta_d|z_d) q(z_d|\rho,\omega^d)\\
    &=\prod_d q(\theta_d|z_d)\prod_i q(z_{d,i} = t | \rho_t, \omega^d)
\end{split}
\end{equation}
\subsection{Variational objective}
With the definition of the variational posterior in Equation \eqref{q_definition}, we can write the variational objective function. Since all the documents in a corpus are generated independently of each other, we derive the objective function for one document and the sum of all such functions as the final objective function of the corpus. The ELBO of document $d$ is defined as:
\begin{equation}\label{elbo_temp}
    \begin{split}
        \mathcal{L}_d  &= \mathbb{E}_{q(\theta_d,z_d|\rho,\omega^d)}[\log p(w_d,\theta_d,z_d)|\alpha,\beta) - \log q(\theta_d,z_d|\rho,\omega^d)]\\
        &= \mathbb{E}_{q(\theta_d,z_d|\rho,\omega^d)}[\log p(\theta_d|\alpha) + \log p(z_d|\theta_d) + \log p(w_d|z_d,\beta)
        - \log q(\theta_d|z_d) - \log q(z_d|\rho,\omega^d)]
        \\
        &= \mathbb{E}_{q(z_d|\rho,\omega^d)}[-\log q(z_d|\rho,\omega^d) - KL_{Dir}(q(\theta_d|z_d)||p(\theta_d|\alpha)) + \log p(w_d|z_d,\beta) + \mathbb{E}_{q(\theta_d|z_d)}[\log p(z_d|\theta_d)]]
    \end{split}
\end{equation}

Combining Equation \eqref{word2topic_gumbel}, \eqref{eta} and \eqref{elbo_temp} we have:
\begin{equation}
\label{elbo}
    \mathcal{L}_d = - \sum_i w_{d,i}(\sum_t \mu_{d,i}^{t}\log\mu_{d,i}^{t}) + \mathbb{E}_{q(z_d|\rho,\omega^d)}[-KL_{Dir}(q(\theta_d|z_d)||p(\theta_d|\alpha)) + \sum_i w_{d,i}\log \beta^{z_{d,i}}_{i} + \mathbb{E}_{q(\theta_d|z_d)}[\sum_t \eta_{d,t}\log \theta_{d,t}]]
\end{equation}
where:
\begin{description}
  \item[$\bullet$] Each $w_{d,i}$ indicates the count of word $i$ in document $d$.
  \item[$\bullet$] Each $z_{d,i}$ denotes the topic assignment of word $i$ in document $d$ and $z_d=[z_{d,0},z_{d,1},...,z_{d,V-1}]$.
  \item[$\bullet$] $\omega^d \in \mathbb{R}^{V \times D}$ is the matrix of all word embeddings of document $d$.
  \item[$\bullet$] $\rho \in \mathbb{R}^{K \times D}$ is the matrix of all the topic vectors.
  \item[$\bullet$] $\alpha$ represents the parameters of the Dirichlet prior in the LDA model.
  \item[$\bullet$] $\beta \in \mathbb{R}^{K \times V}$ is the matrix containing the parameters of all the topic-to-word distributions.
  \item[$\bullet$] Each $\beta^{t}_{i}$ is the probability that topic $t$ generates word $i$. 
  \item[$\bullet$] $KL_{Dir}$ is the KL-divergence between two Dirichlet distributions, which can be calculated analytically.
\end{description}

Here, we treat each document $d$ as a bag-of-words following the standard practice in training LDA using variational inference.

To optimize Equation \eqref{elbo}, we have to use the RT twice: one for the word-to-topic distribution $q(z_d|\rho,\omega^d)$ and one for the document-to-topic distribution $q(\theta_d|z_d)$. For the word-to-topic distribution, a categorical distribution, we use the RT introduced in \cite{jang_categorical_2016,maddison_concrete_2016}, and for the document-to-topic distribution, a Dirichlet distribution, we use the RT introduced in \cite{figurnov_implicit_2018}.

\subsection{Neural network architecture}
\begin{figure}
  \centering
  \includegraphics[width=0.7\textwidth]{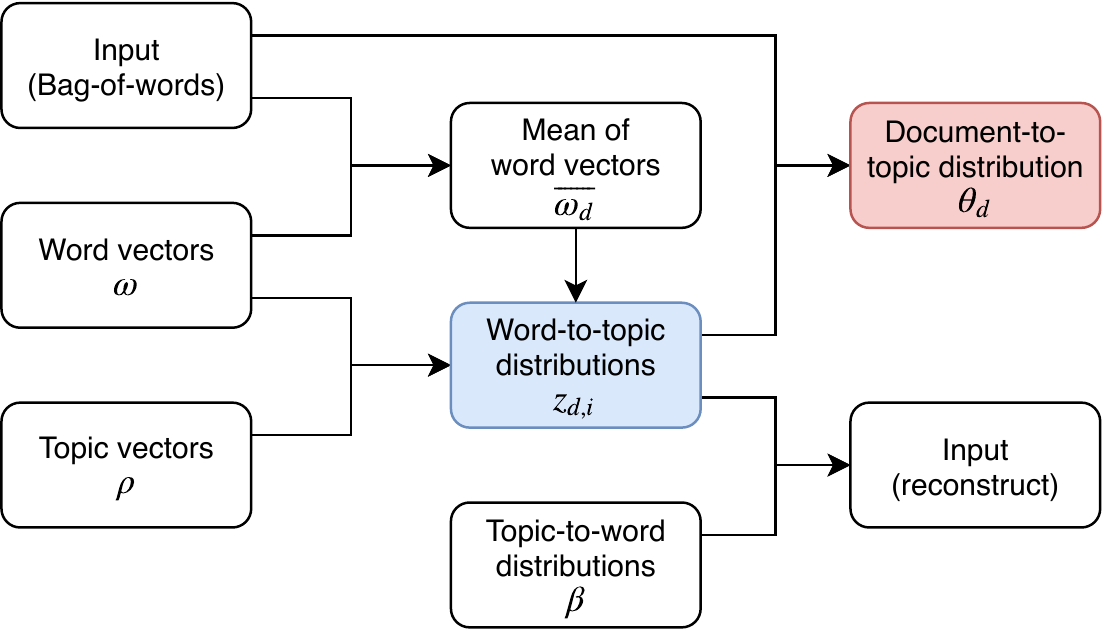}
  \caption{An overview of the neural network architecture of the \modelAbbr{}.}
  \label{fig:vae_nn_overview}
\end{figure}
\begin{figure}
  \centering
  \includegraphics[width=.9\textwidth]{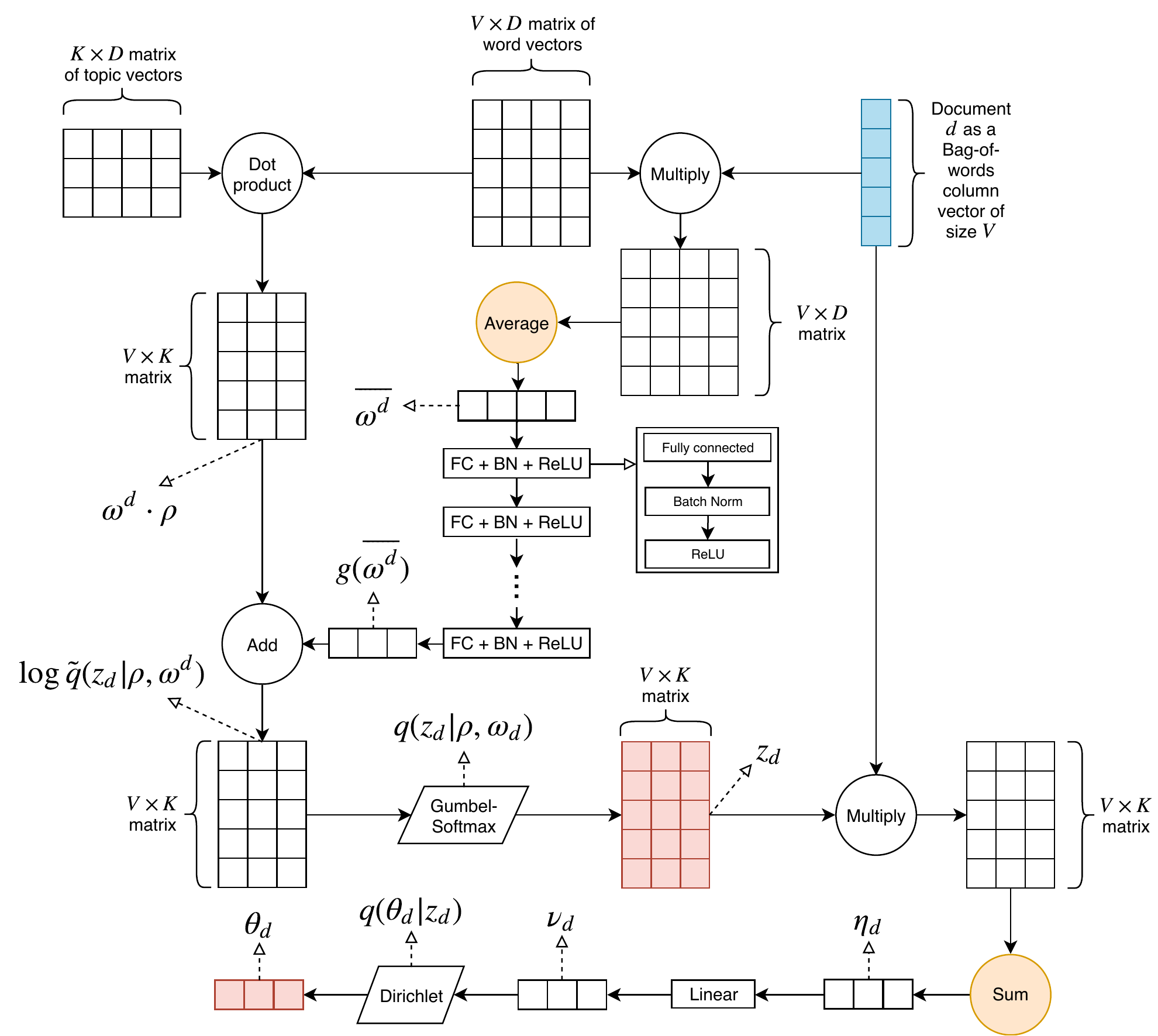}
  \caption{The neural network representation of the posterior distribution $q(\theta,z|\rho,\omega)$. Each dotted line originated from a vector and points to the corresponding factor in the formulas introduced in Section \ref{distribution_q}. Each circle depicts a simple operation, and those with an orange background are operations performed on elements across the first dimension of a matrix. Red vectors are samples drawn from a distribution. ReLU is the function $f(x)=max(0, x)$. Linear is the function $f(x)=ax+b$ where $a$ and $b$ are scalars. A parallelogram depicts a distribution from which we draw samples. $K$ is the number of topics, $D$ is the size of the word embeddings and $V$ is the vocabulary size. Here, $K=3$, $D=4$ and $V=5$. Blue vectors indicate the input of the model.}
  \label{fig:vae_nn}
\end{figure}

From the equations presented in Section \ref{distribution_q}, it is straightforward to translate the variational distribution to a neural network representation. An overview of the architecture is illustrated in Figure \ref{fig:vae_nn_overview}, and the direct mapping between the formula and the architecture is depicted in Figure \ref{fig:vae_nn}.

As seen in the figure, our model receives the embeddings of words within a document as its inputs. In practice, these embeddings are parts of the model's parameters, which gives us the ability to further fine-tune the word embeddings to suit the dataset, should we choose to do so.

Similar to \cite{srivastava_autoencoding_2017}, we found the usage of batch normalization (BN) to be essential to achieving the desire performance of the model, as training without BN frequently leads to \emph{component collapsing}. This phenomenon happens because early in the training process, the KL-divergence between two Dirichlet distributions dominates the loss term, which leads to the model converging to a local minimum where most of the components in the posterior are inactive; hence, the resulting topics are all similar to each other. Moreover, in addition to adding BN between fully connected layers as in Figure \ref{fig:vae_nn}, we also utilize BN in the calculation of the $\beta$ term in Equation \eqref{elbo}:
\begin{align}
    \beta = \texttt{Softmax}(\texttt{Transpose}(\texttt{Batch-Norm}(\texttt{Transpose}(\Tilde{\beta}))))
\end{align}
where $\Tilde{\beta} \in \mathbb{R}^{K \times V}$. Here, we apply BN so that for each topic, the non-normalized probability $\Tilde{\beta}_{t}$ will have a stable mean and variance, which we found necessary for convergence. We hypothesize that such stability provided by BN helps all the topic-to-word distributions to learn at the same rate, i.e., to have gradients with similar magnitudes, which makes the model converge to a more favorable local minimum and avoid component collapsing. Indeed, when measuring the gradients of the topic-to-word distributions, we found the model without BN at the $\beta$ matrix experiencing a divergence - where dominating topics have larger gradients, while other topics' gradients approach zero - and the model with BN having its corresponding gradients stay close to each other. A detailed study will be included in Section \ref{section:bn_effect}.

\section{Experiments and Results}
In this section, we evaluate the \emph{\modelAbbr{}} by measuring its topic coherence and its performance on the document clustering task. 
Topic coherence indicates how related the words that are assigned to each topic are, which closely resembles how humans evaluate a topic model.
The document clustering task directly compares the clustering results to the ground truth labels by considering each class as a cluster.
To demonstrate the advantages of our model, we compare its performance and runtime against the original LDA model and other topic models that utilize word embeddings. To study the significance of word embeddings with respect to the topic model's quality, we compare the results on different sets of pretrained word embeddings. Since our focus is on performing a topic model on a corpus of microtexts, we conduct our experiments mostly on datasets with the average text length smaller than 20 words. We also present the hyperparameters and the training process required for the \emph{\modelAbbr{}} to reach its optimal performance. Finally, we present a detailed study regarding the importance of BN to the model's performance.
Note that for each experiment in the subsequent sections, we ran it $10$ times and reported the average score.

\subsection{Experimental Setup}
\subsubsection{Pretrained word embeddings}
We use two state-of-the-art sets of pretrained word embeddings in our experiments:
\begin{itemize}
    \item Google word vectors\footnote{https://code.google.com/archive/p/word2vec/}: these are 300-dimensional embeddings trained on a subset of the Google News corpus that contains approximately 100 billions words using the Word2Vec framework \citep{mikolov_efficient_2013,NIPS2013_5021}. We denote this set of vectors as \textbf{w2v}.
    \item Stanford word vectors\footnote{https://nlp.stanford.edu/projects/glove/}: these are 300-dimensional embeddings trained on the Common Crawl dataset that contains 42 billions tokens using the GloVe method \citep{pennington_glove:_2014}. We denote this set of vectors as \textbf{glove}.
\end{itemize}
\subsubsection{Datasets}
\begin{table}
\centering
\caption{Statistics of the datasets. \#g: number of ground truth labels. \#d: number of documents. \#w/d: the average length of a document. V: the size of the vocabulary.}
\label{table:dataset}
\begin{tabular}{lllll}
\toprule
\textbf{Dataset}      & \textbf{\#g} & \textbf{\#d}   & \textbf{\#w/d} & \textbf{V}     \\
\midrule
N20          & 20  & 18820 & 103.3 & 19572 \\
N20short     & 20  & 1794  & 13.6  & 6377  \\
N20small     & 20  & 400   & 88    & 8157  \\
TMN          & 7   & 32597 & 18.3  & 13428 \\
TMNtitle     & 7   & 32503 & 4.9   & 6347  \\
Web Snippets & 8   & 12335 & 14.5  & 7314  \\
Twitter      & 4   & 2546  & 4.9   & 1402 \\
\bottomrule
\end{tabular}
\end{table}
We use four datasets: the 20-Newsgroups dataset, the TagMyNews dataset \citep{vitale_classification_2012}, the Sander Twitter corpus and the Web Snippets dataset \citep{phan_hidden_2011} to evaluate our model.

For the 20-Newsgroups and TagMyNews datasets, we use the preprocessed version and their derivation provided by \cite{nguyen_improving_2015}. These include the full version of the 20-Newsgroups dataset and the TagMyNews dataset, denoted \textbf{N20} and \textbf{TMN} respectively; the \textbf{N20short} dataset contains all the documents from the N20 dataset having less than 21 words; the \textbf{N20small} dataset is balanced and contains 400 randomly selected documents from the original N20 dataset; and the \textbf{TMNtitle} dataset consists of only news titles from the TMN dataset.

For the Sander Twitter corpus\footnote{http://www.sananalytics.com/lab/index.php}, we download the 5513 tweets using their Tweet IDs. There are 400 non-downloadable tweets. We closely follow the preprocessing method for this dataset as presented in \cite{nguyen_improving_2015}. After the preprocessing, there are 2546 remaining tweets.

For the Web Snippets\footnote{http://jwebpro.sourceforge.net/data-web-snippets.tar.gz}
dataset, we remove stop words using the list of stop words from the Stanford CoreNLP\footnote{https://github.com/stanfordnlp/CoreNLP},
as well as any words that are not contained in the Stanford and Google pretrained vectors. We also eliminate words that appear less than 3 times in the corpus. Finally, we remove any document whose length is zero after the preprocessing.

Table \ref{table:dataset} summarizes the statistics of all the datasets.

\subsubsection{Baseline models}
We use the following models as baselines in our experiments:
\begin{itemize}
    \item LDA: the original LDA model introduced in \cite{blei_latent_2003}. Here, we consider using Gibbs sampling \citep{griffiths_finding_2004} as the inference method.
    \item LFLDA: a model introduced in \cite{nguyen_improving_2015} that incorporates word vectors to improve the topic model result on short text by using a mixture of the original Dirichlet multinomial and a latent feature component as the topic-to-word distribution. For the experiments, we use the code provided by the authors\footnote{https://github.com/datquocnguyen/LFTM} and the recommended settings in the paper, which includes setting the mixture weight $\lambda$ to $0.6$, the Dirichlet prior $\alpha$ and $\beta$ to $0.1$ and $0.01$, respectively, the number of initial sampling iterations to $1500$ and the number of iterations involving word vectors to $500$.
\end{itemize}
\subsubsection{Training process and hyperparameters}
\begin{table}
\centering
\caption{The hyperparameters for each dataset. \#epochs: number of training epochs. BS: batch size. \#burn-in epochs: number of epochs before training $\alpha$. Min $\tau$: minimum temperature.}
\label{hyperparameters}
\begin{tabular}{@{}lllllll@{}}
\toprule
\textbf{Dataset}      &  \textbf{\#epochs} & \textbf{BS} & \textbf{\begin{tabular}[c]{@{}l@{}}\#burn-in\\ epochs\end{tabular}} & \textbf{Min $\tau$} & \textbf{Layers} & \textbf{\begin{tabular}[c]{@{}l@{}}Train word\\ embedding\end{tabular}} \\ 
\midrule
N20          & 64                                                        & 256        & 32                                                                                 & 0.5                                                           & 128,128 & Yes                                                            \\
N20short     & 256                                                       & 256        & 128                                                                                & 0.7                                                           & 128,128 & No                                                             \\
N20small     & 256                                                       & 200        & 128                                                                                & 0.7                                                           & 128     & No                                                             \\
TMN          & 128                                                       & 256        & 64                                                                                 & 0.7                                                           & 128,128 & No                                                             \\
TMNtitle     & 128                                                       & 256        & 64                                                                                 & 0.7                                                           & 128,128 & No                                                             \\
Web Snippets & 64                                                        & 256        & 32                                                                                 & 0.7                                                           & 128,128 & No                                                             \\
Twitter      & 128                                                       & 256        & 64                                                                                 & 0.7                                                           & 128,128 & No                                                             \\
\bottomrule
\end{tabular}
\end{table}
To train the \emph{\modelAbbr{}}, we use the Adam Optimizer \citep{kingma_adam:_2014} with $\beta_1 = 0$ and $\beta_2 = 0.99$. We use a low momentum value since it allows the topic placement of a word to change quickly during the training process. From our experiments, we are able to confirm that using a low momentum allows our model to reach a lower perplexity as well as improve both topic coherence scores and document clustering results. We set the learning rate to $8e-3$ since it allows our model to converge quickly.

When training, we slowly increase the learning rate from $0$ to $8e-3$ while decreasing the temperature $\tau$ of the Gumbel-Softmax from $1$ to a minimum value - which is $0.7$ for small datasets or datasets of microtexts and $0.5$ for large datasets of long texts. This process takes place in the first epoch, which acts as a warm-up period for the model.

We set the initial value of the Dirichlet prior $\alpha$ to $0.1$, and optimize this value after a certain number of epochs, which is usually one-half the total amount of epochs. We find this greatly improves our model performance, especially on the document clustering task, which is not surprising since the study in \cite{wallach_rethinking_2009} confirmed the importance of the Dirichlet prior to the quality of the topic model.

We use a batch size of $200$ for the N20small dataset and of $256$ for other datasets. We choose the number of epochs for each dataset that guarantees the convergence of our model. We use one fully connected layer of 128 units for the N20small dataset and two fully connected layers of 128 units for other datasets.

On large datasets of long texts, such as the N20 dataset, we also allow the training of word embeddings, which resulted in an additional boost in the model performance.

Table \ref{hyperparameters} presents the hyperparameters for each dataset.

\subsection{Topic coherence}
\begin{table}
\centering
\caption{NPMI scores (higher is better) for the N20short, TMN, TMNtitle, Twitter and Web Snippets dataset.}
\label{NPMI score}
\begin{tabular}{@{}lcrrrrr@{}}
\toprule
\textbf{Dataset}                   & \multicolumn{1}{c}{\begin{tabular}[c]{@{}c@{}}\textbf{Number of}\\\textbf{topics}\end{tabular}} & \multicolumn{1}{c}{\begin{tabular}[c]{@{}c@{}}\textbf{LFLDA}\\\textbf{glove}\end{tabular}} & \multicolumn{1}{c}{\textbf{\begin{tabular}[c]{@{}c@{}}\textbf{LFLDA}\\\textbf{w2v}\end{tabular}}} & \multicolumn{1}{c}{\begin{tabular}[c]{@{}c@{}}\textbf{N-VAE}\\\textbf{glove}\end{tabular}} & \multicolumn{1}{c}{\textbf{\begin{tabular}[c]{@{}c@{}}\textbf{N-VAE}\\\textbf{w2v}\end{tabular}}} & \multicolumn{1}{c}{\textbf{LDA}} \\ 
\midrule
\multirow{4}{*}{N20short} & 6                                                                              & -3.70$\pm$0.37                                                            & -3.86$\pm$1.01                                                          & -4.29$\pm$1.30                                                            & \textbf{-3.62$\pm$1.03}                                                 & \textbf{-3.62$\pm$0.74} \\
                          & 20                                                                             & -5.13$\pm$0.66                                                            & \textbf{-4.68$\pm$0.71}                                                 & -5.23$\pm$0.91                                                            & -5.57$\pm$1.00                                                          & -5.18$\pm$0.68          \\
                          & 40                                                                             & -6.42$\pm$0.32                                                            & \textbf{-6.37$\pm$0.41}                                                 & -7.62$\pm$0.64                                                            & -6.95$\pm$0.99                                                          & -6.69$\pm$0.24          \\
                          & 80                                                                             & -8.04$\pm$0.34                                                            & -7.88$\pm$0.20                                                          & -9.01$\pm$0.66                                                            & \textbf{-7.14$\pm$1.05}                                                 & -8.86$\pm$0.28          \\ \midrule
\multirow{4}{*}{TMN}      & 7                                                                              & \textbf{3.83$\pm$1.14}                                                    & 3.60$\pm$0.74                                                           & 3.77$\pm$0.63                                                             & 3.72$\pm$1.44                                                           & 3.47$\pm$0.91           \\
                          & 20                                                                             & 3.74$\pm$0.54                                                             & \textbf{3.97$\pm$0.52}                                                  & 2.35$\pm$1.39                                                             & 3.16$\pm$1.14                                                           & 3.20$\pm$0.70           \\
                          & 40                                                                             & 2.83$\pm$0.41                                                             & \textbf{3.00$\pm$0.55}                                                  & -3.43$\pm$1.27                                                            & -2.34$\pm$1.38                                                          & 2.86$\pm$0.32           \\
                          & 80                                                                             & \textbf{2.01$\pm$0.31}                                                    & 2.00$\pm$0.18                                                           & -7.34$\pm$0.57                                                            & -7.07$\pm$1.18                                                          & 1.61$\pm$0.33           \\ \midrule
\multirow{4}{*}{TMNtitle} & 7                                                                              & -0.44$\pm$0.95                                                            & -0.71$\pm$0.81                                                          & \textbf{-0.17$\pm$0.98}                                                   & -0.43$\pm$1.51                                                          & -0.94$\pm$0.69          \\
                          & 20                                                                             & \textbf{0.10$\pm$0.58}                                                    & -0.02$\pm$0.59                                                          & -3.81$\pm$1.04                                                            & -2.79$\pm$1.76                                                          & 0.01$\pm$0.52           \\
                          & 40                                                                             & -0.39$\pm$0.31                                                            & \textbf{-0.32$\pm$0.28}                                                 & -7.48$\pm$1.74                                                            & -5.46$\pm$0.97                                                          & -0.62$\pm$0.24          \\
                          & 80                                                                             & \textbf{-1.29$\pm$0.10}                                                   & -1.31$\pm$0.34                                                          & -7.50$\pm$0.73                                                            & -5.47$\pm$0.87                                                          & -1.60$\pm$0.28          \\ \midrule
\multirow{4}{*}{Twitter}  & 4                                                                              & -2.50$\pm$0.54                                                            & -2.70$\pm$0.94                                                          & \textbf{-1.99$\pm$1.56}                                                   & -2.06$\pm$1.73                                                          & -3.09$\pm$0.96          \\
                          & 20                                                                             & -6.47$\pm$0.57                                                            & -6.30$\pm$0.40                                                          & -5.52$\pm$0.79                                                            & \textbf{-5.17$\pm$1.03}                                                 & -6.45$\pm$0.53          \\
                          & 40                                                                             & -7.41$\pm$0.19                                                            & -7.48$\pm$0.38                                                          & -6.98$\pm$0.73                                                            & \textbf{-5.43$\pm$0.35}                                                 & -7.92$\pm$0.41          \\
                          & 80                                                                             & -8.74$\pm$0.33                                                            & -8.36$\pm$0.25                                                          & -6.17$\pm$0.58                                                            & \textbf{-6.16$\pm$0.67}                                                 & -9.43$\pm$0.19          \\ \midrule
\multirow{4}{*}{snippets} & 8                                                                              & 0.93$\pm$1.02                                                             & 0.67$\pm$1.23                                                           & \textbf{2.14$\pm$1.01}                                                    & 1.91$\pm$1.07                                                           & 0.98$\pm$0.91           \\
                          & 20                                                                             & 1.62$\pm$0.61                                                             & 1.92$\pm$1.38                                                           & 2.21$\pm$1.32                                                             & \textbf{2.98$\pm$0.75}                                                  & 1.66$\pm$0.74           \\
                  & 40                                                                             & -0.07$\pm$0.48                                                            & 0.18$\pm$0.44                                                           & -3.97$\pm$0.73                                                            & \textbf{0.96$\pm$0.66}                                                  & -0.19$\pm$0.77          \\                          & 80                                                                             & -1.54$\pm$0.34                                                            & -1.54$\pm$0.19                                                          & -8.57$\pm$1.30                                                            & \textbf{-1.39$\pm$0.59}                                                 & -1.90$\pm$0.43          \\
\bottomrule
\end{tabular}
\end{table}
We examine the quality of each topic produced by our model via measuring how semantically coherent its top words are.

Based on the survey done in \cite{roder_exploring_2015}, we use normalized pointwise mutual information (NPMI) as the metric for quantitative analysis of topic coherence. This metric is introduced in \cite{bouma_normalized_2009} and is proven empirically to have a strong correlation with human evaluation. The NPMI is defined as:
\begin{equation}\label{eq:npmi}
    \text{NPMI}(w_i, w_j) = \sum_{i=0}^{N-2}\sum_{j=i+1}^{N-1} -\frac{\log \frac{P(w_i,w_j)}{P(w_i)P(w_j)}}{\log P(w_i, w_j)}
\end{equation}
where the probabilities are collected using a sliding window of 10 words on an external corpus. For each topic, its NPMI score is calculated using its top-15 words.
We use \emph{Palmetto}\footnote{ https://github.com/dice-group/Palmetto}, the tool provided by the authors of \cite{roder_exploring_2015} to measure topic coherence using Wikipedia as the external corpus. For each dataset, we calculate the coherence of each topic and use the average score of all the topics as the model's coherence.

Table \ref{NPMI score} presents the NPMI scores from various models with different numbers of topics. Since the LFLDA model requires a long time to train, our time and resource limitations only permit us to conduct experiments on $5$ datasets: N20short, TMN, TMNtitle, Twitter and Web Snippets. The table shows that our model produces competitive results, frequently having the highest NPMI scores when combined with Word2Vec. 

\subsection{Document clustering}
\begin{table}
\centering
\caption{NMI scores (higher is better) for the N20, N20small, N20short, TMN, TMNtitle, Twitter and Web Snippets dataset.}
\label{NMI score}
\begin{tabular}{@{}lcrrrrr@{}}
\toprule
\textbf{Dataset}                   & \multicolumn{1}{c}{\begin{tabular}[c]{@{}c@{}}\textbf{Number of}\\\textbf{topics}\end{tabular}} & \multicolumn{1}{c}{\begin{tabular}[c]{@{}c@{}}\textbf{LFLDA}\\\textbf{glove}\end{tabular}} & \multicolumn{1}{c}{\textbf{\begin{tabular}[c]{@{}c@{}}\textbf{LFLDA}\\\textbf{w2v}\end{tabular}}} & \multicolumn{1}{c}{\begin{tabular}[c]{@{}c@{}}\textbf{N-VAE}\\\textbf{glove}\end{tabular}} & \multicolumn{1}{c}{\textbf{\begin{tabular}[c]{@{}c@{}}\textbf{N-VAE}\\\textbf{w2v}\end{tabular}}} & \multicolumn{1}{c}{\textbf{LDA}} \\ \midrule
\multirow{4}{*}{N20}      & 6                                                                              & \textbf{0.522$\pm$0.003}                                                  & 0.500$\pm$0.008                                                         & 0.519$\pm$0.014                                                           & 0.513$\pm$0.019                                                         & 0.516$\pm$0.009         \\
                          & 20                                                                             & 0.596$\pm$0.012                                                           & 0.563$\pm$0.009                                                         & \textbf{0.632$\pm$0.014}                                                  & 0.620$\pm$0.012                                                         & 0.582$\pm$0.009         \\
                          & 40                                                                             & 0.550$\pm$0.010                                                           & 0.535$\pm$0.008                                                         & \textbf{0.624$\pm$0.012}                                                  & 0.617$\pm$0.012                                                         & 0.557$\pm$0.007         \\
                          & 80                                                                             & 0.507$\pm$0.003                                                           & 0.505$\pm$0.004                                                         & \textbf{0.613$\pm$0.004}                                                  & 0.600$\pm$0.013                                                         & 0.515$\pm$0.003         \\ \midrule
\multirow{4}{*}{N20small} & 6                                                                              & 0.436$\pm$0.019                                                           & 0.406$\pm$0.023                                                         & \textbf{0.437$\pm$0.019}                                                  & 0.403$\pm$0.022                                                         & 0.376$\pm$0.016         \\
                          & 20                                                                             & 0.504$\pm$0.020                                                           & 0.519$\pm$0.014                                                         & \textbf{0.533$\pm$0.023}                                                  & 0.492$\pm$0.017                                                         & 0.474$\pm$0.013         \\
                          & 40                                                                             & 0.527$\pm$0.013                                                           & 0.548$\pm$0.017                                                         & \textbf{0.560$\pm$0.011}                                                  & 0.514$\pm$0.012                                                         & 0.513$\pm$0.009         \\
                          & 80                                                                             & 0.576$\pm$0.006                                                           & \textbf{0.585$\pm$0.009}                                                & \textbf{0.585$\pm$0.012}                                                  & 0.554$\pm$0.017                                                         & 0.563$\pm$0.008         \\ \midrule
\multirow{4}{*}{N20short} & 6                                                                              & 0.290$\pm$0.011                                                           & 0.269$\pm$0.012                                                         & \textbf{0.301$\pm$0.025}                                                  & 0.252$\pm$0.032                                                         & 0.224$\pm$0.012         \\
                          & 20                                                                             & 0.294$\pm$0.007                                                           & 0.296$\pm$0.012                                                         & \textbf{0.353$\pm$0.013}                                                  & 0.302$\pm$0.017                                                         & 0.248$\pm$0.009         \\
                          & 40                                                                             & 0.299$\pm$0.009                                                           & 0.309$\pm$0.008                                                         & \textbf{0.375$\pm$0.015}                                                  & 0.311$\pm$0.015                                                         & 0.268$\pm$0.010         \\
                          & 80                                                                             & 0.338$\pm$0.008                                                           & 0.340$\pm$0.005                                                         & \textbf{0.367$\pm$0.011}                                                  & 0.294$\pm$0.018                                                         & 0.308$\pm$0.009         \\ \midrule
\multirow{4}{*}{TMN}      & 7                                                                              & 0.448$\pm$0.017                                                           & 0.446$\pm$0.014                                                         & \textbf{0.487$\pm$0.031}                                                  & 0.467$\pm$0.034                                                         & 0.429$\pm$0.026         \\
                          & 20                                                                             & 0.403$\pm$0.004                                                           & 0.399$\pm$0.006                                                         & \textbf{0.467$\pm$0.011}                                                  & 0.457$\pm$0.011                                                         & 0.400$\pm$0.009         \\
                          & 40                                                                             & 0.356$\pm$0.004                                                           & 0.355$\pm$0.005                                                         & \textbf{0.464$\pm$0.013}                                                  & 0.452$\pm$0.013                                                         & 0.354$\pm$0.004         \\
                          & 80                                                                             & 0.324$\pm$0.004                                                           & 0.325$\pm$0.003                                                         & \textbf{0.474$\pm$0.015}                                                  & 0.447$\pm$0.031                                                         & 0.319$\pm$0.006         \\ \midrule
\multirow{4}{*}{TMNtitle} & 7                                                                              & 0.322$\pm$0.015                                                           & 0.321$\pm$0.012                                                         & \textbf{0.400$\pm$0.026}                                                  & 0.367$\pm$0.028                                                         & 0.317$\pm$0.013         \\
                          & 20                                                                             & 0.280$\pm$0.004                                                           & 0.279$\pm$0.006                                                         & \textbf{0.395$\pm$0.020}                                                  & 0.361$\pm$0.025                                                         & 0.269$\pm$0.005         \\
                          & 40                                                                             & 0.235$\pm$0.006                                                           & 0.239$\pm$0.005                                                         & \textbf{0.393$\pm$0.028}                                                  & 0.362$\pm$0.028                                                         & 0.227$\pm$0.006         \\
                          & 80                                                                             & 0.209$\pm$0.003                                                           & 0.210$\pm$0.002                                                         & \textbf{0.372$\pm$0.017}                                                  & 0.341$\pm$0.018                                                         & 0.196$\pm$0.003         \\ \midrule
\multirow{4}{*}{Twitter}  & 4                                                                              & 0.260$\pm$0.022                                                           & 0.263$\pm$0.031                                                         & 0.262$\pm$0.032                                                           & \textbf{0.274$\pm$0.031}                                                & 0.234$\pm$0.021         \\
                          & 20                                                                             & 0.184$\pm$0.005                                                           & 0.192$\pm$0.007                                                         & 0.249$\pm$0.011                                                           & \textbf{0.251$\pm$0.025}                                                & 0.178$\pm$0.008         \\
                          & 40                                                                             & 0.175$\pm$0.006                                                           & 0.180$\pm$0.004                                                         & 0.256$\pm$0.018                                                           & \textbf{0.276$\pm$0.019}                                                & 0.162$\pm$0.007         \\
                          & 80                                                                             & 0.163$\pm$0.006                                                           & 0.167$\pm$0.004                                                         & 0.240$\pm$0.027                                                           & \textbf{0.267$\pm$0.027}                                                & 0.156$\pm$0.007         \\ \midrule
\multirow{4}{*}{snippets} & 8                                                                              & 0.523$\pm$0.035                                                           & 0.516$\pm$0.032                                                         & 0.628$\pm$0.034                                                           & \textbf{0.630$\pm$0.033}                                                & 0.490$\pm$0.039         \\
                          & 20                                                                             & 0.433$\pm$0.011                                                           & 0.430$\pm$0.015                                                         & 0.590$\pm$0.014                                                           & \textbf{0.597$\pm$0.020}                                                & 0.424$\pm$0.017         \\
                          & 40                                                                             & 0.372$\pm$0.006                                                           & 0.381$\pm$0.007                                                         & 0.573$\pm$0.010                                                           & \textbf{0.574$\pm$0.015}                                                & 0.364$\pm$0.008         \\
                          & 80                                                                             & 0.346$\pm$0.005                                                           & 0.349$\pm$0.007                                                         & \textbf{0.587$\pm$0.021}                                                  & 0.579$\pm$0.018                                                         & 0.336$\pm$0.011         \\ 
\bottomrule
\end{tabular}
\end{table}
\begin{table}
\centering
\caption{Purity scores (higher is better) for the N20, N20small, N20short, TMN, TMNtitle, Twitter and Web Snippets dataset.}
\label{Purity score}
\begin{tabular}{@{}lcrrrrr@{}}
\toprule
\textbf{Dataset}                   & \multicolumn{1}{c}{\begin{tabular}[c]{@{}c@{}}\textbf{Number of}\\\textbf{topics}\end{tabular}} & \multicolumn{1}{c}{\begin{tabular}[c]{@{}c@{}}\textbf{LFLDA}\\\textbf{glove}\end{tabular}} & \multicolumn{1}{c}{\textbf{\begin{tabular}[c]{@{}c@{}}\textbf{LFLDA}\\\textbf{w2v}\end{tabular}}} & \multicolumn{1}{c}{\begin{tabular}[c]{@{}c@{}}\textbf{N-VAE}\\\textbf{glove}\end{tabular}} & \multicolumn{1}{c}{\textbf{\begin{tabular}[c]{@{}c@{}}\textbf{N-VAE}\\\textbf{w2v}\end{tabular}}} & \multicolumn{1}{c}{\textbf{LDA}} \\ \midrule
\multirow{4}{*}{N20}      & 6                                                                              & \textbf{0.295$\pm$0.001}                                                  & 0.291$\pm$0.002                                                         & 0.293$\pm$0.004                                                           & 0.288$\pm$0.011                                                         & 0.293$\pm$0.002         \\
                          & 20                                                                             & \textbf{0.604$\pm$0.031}                                                  & 0.569$\pm$0.021                                                         & 0.602$\pm$0.032                                          & 0.584$\pm$0.028                                                         & 0.573$\pm$0.019         \\
                          & 40                                                                             & 0.632$\pm$0.017                                                           & 0.616$\pm$0.017                                                         & \textbf{0.652$\pm$0.024}                                                  & 0.634$\pm$0.027                                                         & 0.639$\pm$0.017         \\
                          & 80                                                                             & 0.638$\pm$0.007                                                           & 0.638$\pm$0.006                                                         & \textbf{0.664$\pm$0.020}                                                  & 0.652$\pm$0.018                                                         & 0.646$\pm$0.005         \\ \midrule
\multirow{4}{*}{N20small} & 6                                                                              & 0.235$\pm$0.008                                                           & 0.229$\pm$0.005                                                         & \textbf{0.252$\pm$0.008}                                                  & 0.237$\pm$0.015                                                         & 0.232$\pm$0.011         \\
                          & 20                                                                             & 0.427$\pm$0.022                                                           & 0.439$\pm$0.015                                                         & \textbf{0.451$\pm$0.023}                                                  & 0.408$\pm$0.025                                                         & 0.408$\pm$0.017         \\
                          & 40                                                                             & 0.492$\pm$0.022                                                           & 0.516$\pm$0.024                                                         & \textbf{0.524$\pm$0.026}                                                  & 0.465$\pm$0.024                                                         & 0.477$\pm$0.015         \\
                          & 80                                                                             & 0.579$\pm$0.011                                                           & \textbf{0.595$\pm$0.016}                                                & 0.586$\pm$0.017                                                           & 0.543$\pm$0.028                                                         & 0.559$\pm$0.018         \\ \midrule
\multirow{4}{*}{N20short} & 6                                                                              & 0.287$\pm$0.017                                                           & 0.278$\pm$0.007                                                         & \textbf{0.297$\pm$0.028}                                                  & 0.269$\pm$0.025                                                         & 0.265$\pm$0.020         \\
                          & 20                                                                             & 0.358$\pm$0.011                                                           & 0.366$\pm$0.010                                                         & \textbf{0.375$\pm$0.022}                                                  & 0.330$\pm$0.026                                                         & 0.320$\pm$0.017         \\
                          & 40                                                                             & 0.380$\pm$0.015                                                           & 0.395$\pm$0.013                                                         & \textbf{0.400$\pm$0.028}                                                  & 0.345$\pm$0.017                                                         & 0.353$\pm$0.014         \\
                          & 80                                                                             & 0.425$\pm$0.015                                                           & \textbf{0.429$\pm$0.009}                                                & 0.373$\pm$0.014                                                           & 0.326$\pm$0.021                                                         & 0.395$\pm$0.009         \\ \midrule
\multirow{4}{*}{TMN}      & 7                                                                              & 0.658$\pm$0.034                                                           & 0.658$\pm$0.020                                                         & \textbf{0.675$\pm$0.040}                                                  & 0.662$\pm$0.040                                                         & 0.645$\pm$0.037         \\
                          & 20                                                                             & 0.722$\pm$0.007                                                           & 0.716$\pm$0.012                                                         & 0.754$\pm$0.016                                                           & \textbf{0.758$\pm$0.015}                                                & 0.725$\pm$0.013         \\
                          & 40                                                                             & 0.719$\pm$0.008                                                           & 0.720$\pm$0.008                                                         & \textbf{0.766$\pm$0.014}                                                  & 0.755$\pm$0.011                                                         & 0.719$\pm$0.007         \\
                          & 80                                                                             & 0.725$\pm$0.006                                                           & 0.725$\pm$0.004                                                         & \textbf{0.741$\pm$0.025}                                                  & 0.701$\pm$0.042                                                         & 0.716$\pm$0.008         \\ \midrule
\multirow{4}{*}{TMNtitle} & 7                                                                              & 0.584$\pm$0.026                                                           & 0.579$\pm$0.020                                                         & \textbf{0.634$\pm$0.030}                                                  & 0.603$\pm$0.038                                                         & 0.581$\pm$0.016         \\
                          & 20                                                                             & 0.623$\pm$0.012                                                           & 0.619$\pm$0.015                                                         & \textbf{0.662$\pm$0.029}                                                  & 0.629$\pm$0.026                                                         & 0.610$\pm$0.010         \\
                          & 40                                                                             & 0.600$\pm$0.008                                                           & 0.611$\pm$0.007                                                         & \textbf{0.654$\pm$0.042}                                                  & 0.622$\pm$0.041                                                         & 0.593$\pm$0.011         \\
                          & 80                                                                             & 0.601$\pm$0.004                                                           & 0.598$\pm$0.004                                                         & \textbf{0.632$\pm$0.025}                                                  & 0.588$\pm$0.035                                                         & 0.582$\pm$0.005         \\ \midrule
\multirow{4}{*}{Twitter}  & 4                                                                              & 0.606$\pm$0.030                                                           & \textbf{0.615$\pm$0.039}                                                & 0.594$\pm$0.029                                                           & 0.603$\pm$0.034                                                         & 0.592$\pm$0.026         \\
                          & 20                                                                             & 0.626$\pm$0.014                                                           & 0.637$\pm$0.013                                                         & \textbf{0.665$\pm$0.016}                                                  & 0.651$\pm$0.025                                                         & 0.611$\pm$0.015         \\
                          & 40                                                                             & 0.633$\pm$0.009                                                           & 0.640$\pm$0.008                                                         & 0.661$\pm$0.024                                                           & \textbf{0.669$\pm$0.029}                                                & 0.616$\pm$0.009         \\
                          & 80                                                                             & 0.630$\pm$0.007                                                           & 0.636$\pm$0.006                                                         & 0.654$\pm$0.040                                                           & \textbf{0.659$\pm$0.037}                                                & 0.624$\pm$0.011         \\ \midrule
\multirow{4}{*}{snippets} & 8                                                                              & 0.741$\pm$0.036                                                           & 0.730$\pm$0.045                                                         & 0.790$\pm$0.044                                                           & \textbf{0.796$\pm$0.041}                                                & 0.707$\pm$0.042         \\
                          & 20                                                                             & 0.725$\pm$0.013                                                           & 0.724$\pm$0.015                                                         & 0.849$\pm$0.020                                                           & \textbf{0.853$\pm$0.023}                                                & 0.721$\pm$0.022         \\
                          & 40                                                                             & 0.696$\pm$0.009                                                           & 0.709$\pm$0.010                                                         & 0.855$\pm$0.017                                                           & \textbf{0.856$\pm$0.015}                                                & 0.686$\pm$0.012         \\
                          & 80                                                                             & 0.701$\pm$0.010                                                           & 0.703$\pm$0.012                                                         & \textbf{0.866$\pm$0.016}                                                  & 0.847$\pm$0.015                                                         & 0.685$\pm$0.015         \\ 
\bottomrule
\end{tabular}
\end{table}
We measure to what extent the clustering result of the topic model agrees with the ground truth label. After calculating the topic proportion for each document using a topic model, we consider a document to belong to the topic with the highest probability in that document. We then calculate the similarity between clusters produced by the topic model and the ground truth label using two metrics: purity and normalized mutual information (NMI). For details on these two metrics, please refer to \cite{manning_2008}, Section 16.3.

Table \ref{NMI score} and \ref{Purity score}, respectively presents the NMI and purity scores from various models with different numbers of topics. Note that for the N20 and N20small dataset, we use the results of the LFLDA-glove and LFLDA-w2v models reported in \cite{nguyen_improving_2015}. The table shows that the \emph{\modelAbbr{}} outperforms other models most of the time, and using GloVe will frequently lead to better scores than using Word2Vec. Even on the N20 dataset, the additional information from the word vector is still useful and allows our model to reach a much better score than the original LDA model.

\subsection{Runtime}
\begin{table}
\centering
\caption{Runtime of the LFLDA and the \modelAbbr{} on the N20short dataset with various numbers of topics.}
\label{Runtime}
\begin{tabular}{@{}lccc@{}}
\toprule
Dataset  & Number of topics & LFLDA (s) & \modelAbbr{} (s) \\
\midrule
N20short & 6            & 1266      & 40               \\
         & 20           & 2007      & 48               \\
         & 40           & 3247      & 61               \\
         & 80           & 4068      & 73              \\
\bottomrule
\end{tabular}
\end{table}
We compare the runtime between the LFLDA and the \modelAbbr{}. We ran each model on the N20short dataset using a machine with 24 vCPUs and 90 GB of RAM rented on the Google Compute Engine\footnote{ https://cloud.google.com/compute/}. Even though a GPU could be utilized to speed up the training process of our model, we only use the CPU for fairness. The results are shown in table \ref{Runtime}. Overall, our model requires less than $1/30$th the training time of the LFLDA to reach the desired performance. This is the largest advantage of our model, enabling training on a large corpus while requiring only a small amount of time.

\subsection{Effect of Batch Normalization}
\label{section:bn_effect}
\begin{figure}
\begin{subfigure}{.33\textwidth}
  \centering
  \includegraphics[width=\linewidth]{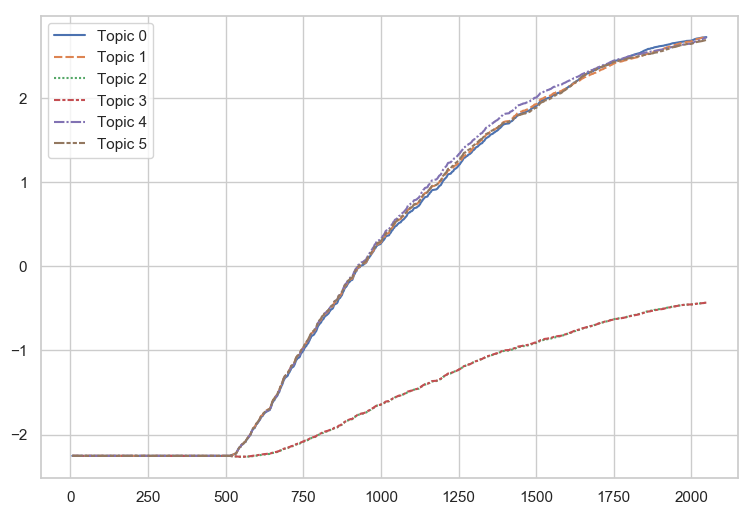}
  \caption{The value of $\alpha$}
  
\end{subfigure}%
\begin{subfigure}{.33\textwidth}
  \centering
  \includegraphics[width=\linewidth]{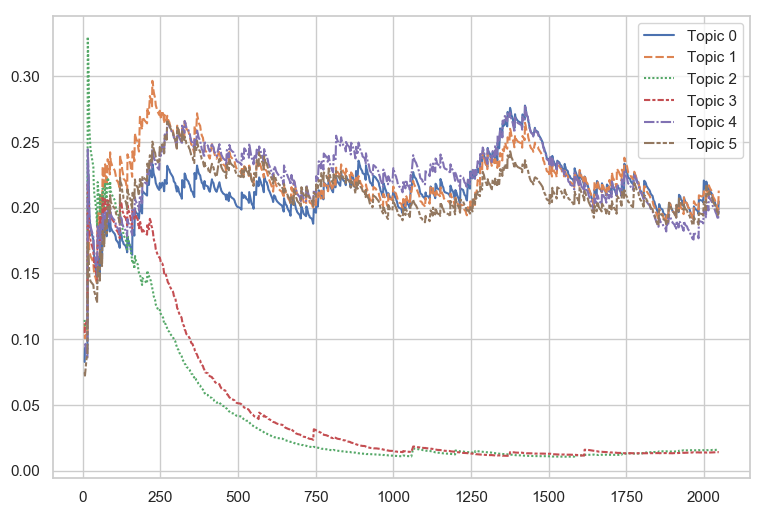}
  \caption{$\beta$'s gradients}
  
\end{subfigure}
\begin{subfigure}{.33\textwidth}
  \centering
  \includegraphics[width=\linewidth]{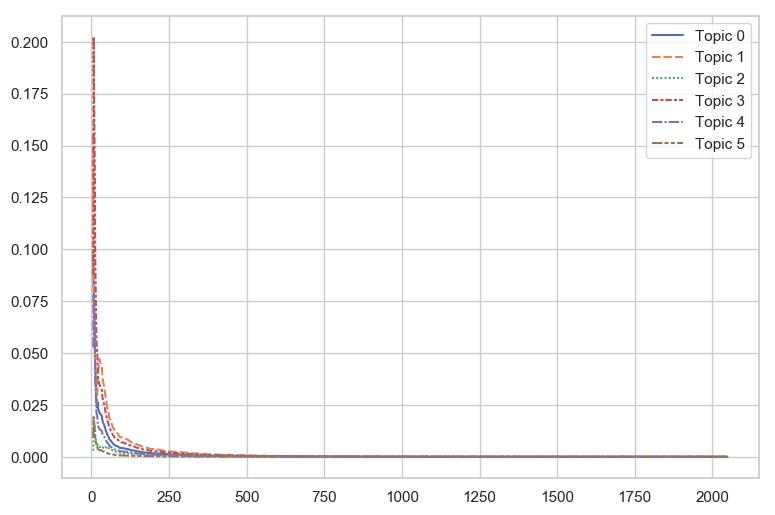}
  \caption{The last FC layer's weights' gradients}
  
\end{subfigure}
\caption{No BN between the FC layers and at the $\beta$ matrix}
\label{bn:no-no}
\end{figure}
\begin{figure}
\begin{subfigure}{.33\textwidth}
  \centering
  \includegraphics[width=\linewidth]{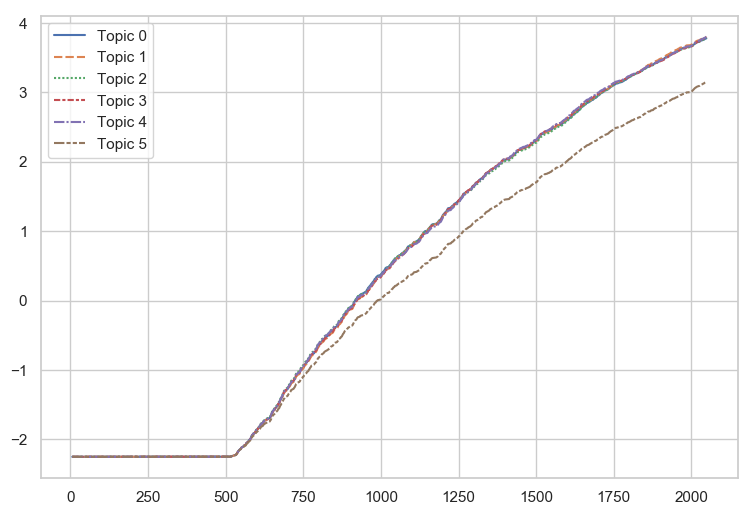}
  \caption{The value of $\alpha$}
  
\end{subfigure}%
\begin{subfigure}{.33\textwidth}
  \centering
  \includegraphics[width=\linewidth]{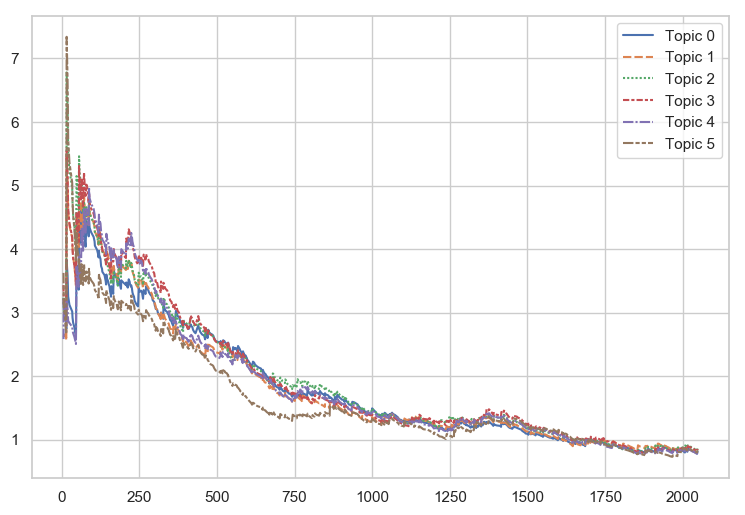}
  \caption{$\beta$'s gradients}
  
\end{subfigure}
\begin{subfigure}{.33\textwidth}
  \centering
  \includegraphics[width=\linewidth]{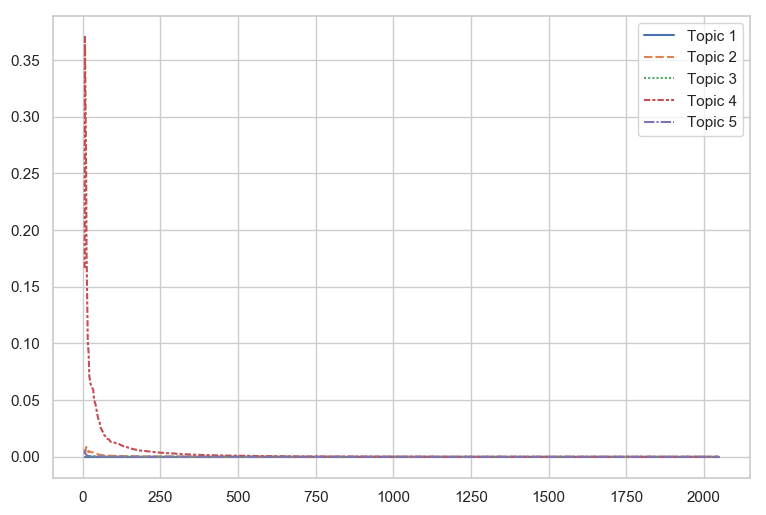}
  \caption{The last FC layer's weights' gradients}
  
\end{subfigure}
\caption{BN only at the $\beta$ matrix and none between the FC layers.}
\label{bn:no-yes}
\end{figure}
\begin{figure}
\begin{subfigure}{.33\textwidth}
  \centering
  \includegraphics[width=\linewidth]{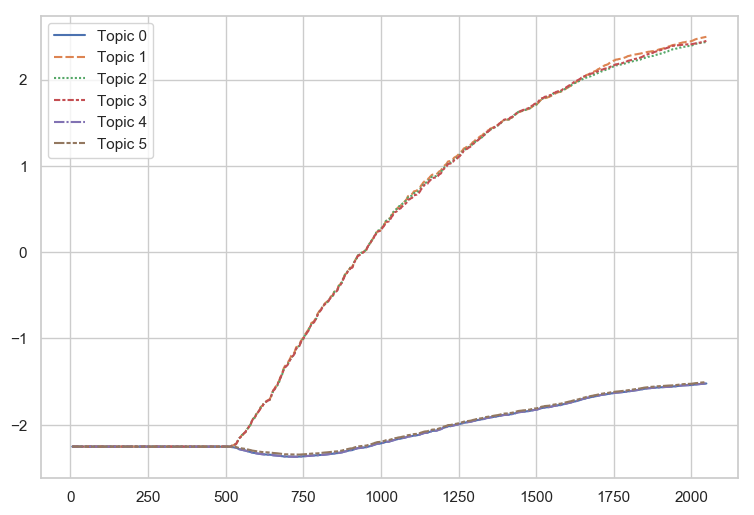}
  \caption{The value of $\alpha$}
  
\end{subfigure}%
\begin{subfigure}{.33\textwidth}
  \centering
  \includegraphics[width=\linewidth]{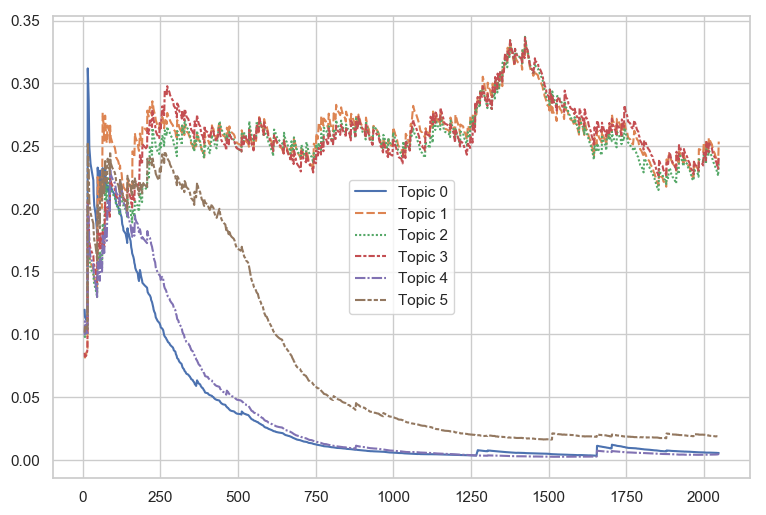}
  \caption{$\beta$'s gradients}
  
\end{subfigure}
\begin{subfigure}{.33\textwidth}
  \centering
  \includegraphics[width=\linewidth]{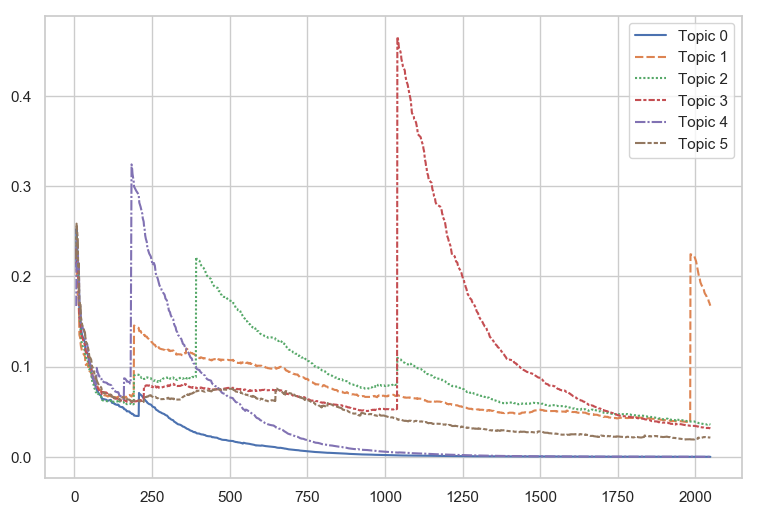}
  \caption{The last FC layer's weights' gradients}
  
\end{subfigure}
\caption{BN only between the FC layers and none at the $\beta$ matrix.}
\label{bn:yes-no}
\end{figure}
\begin{figure}
\begin{subfigure}{.33\textwidth}
  \centering
  \includegraphics[width=\linewidth]{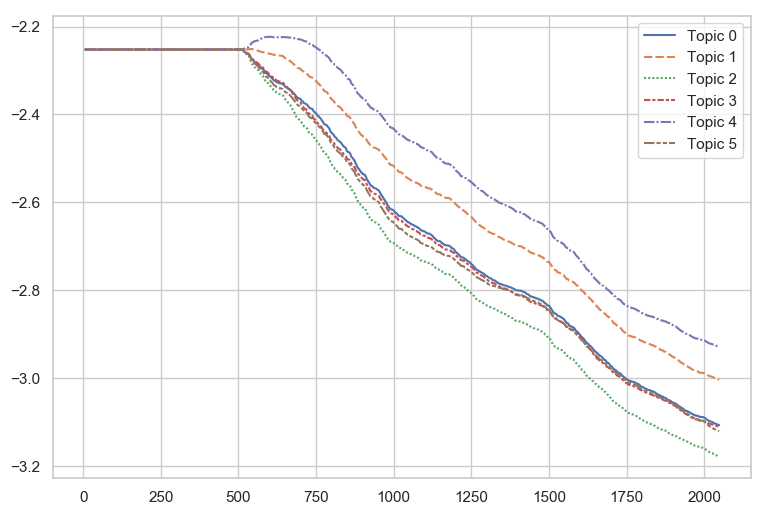}
  \caption{The value of $\alpha$}
  
\end{subfigure}%
\begin{subfigure}{.33\textwidth}
  \centering
  \includegraphics[width=\linewidth]{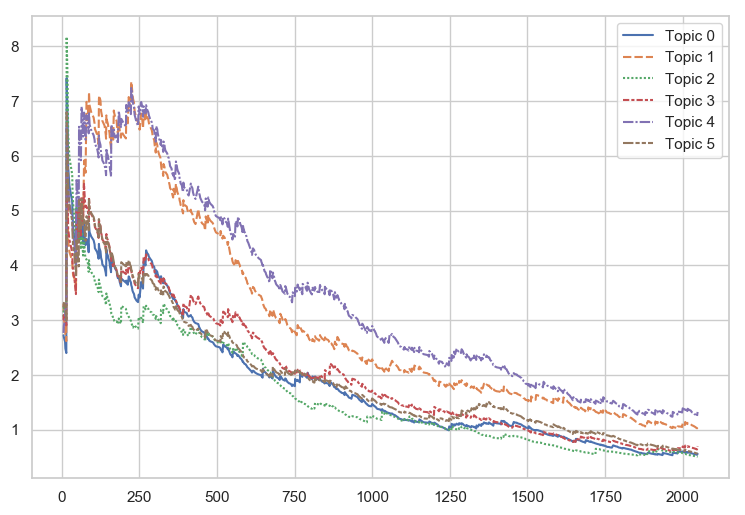}
  \caption{$\beta$'s gradients}
  
\end{subfigure}
\begin{subfigure}{.33\textwidth}
  \centering
  \includegraphics[width=\linewidth]{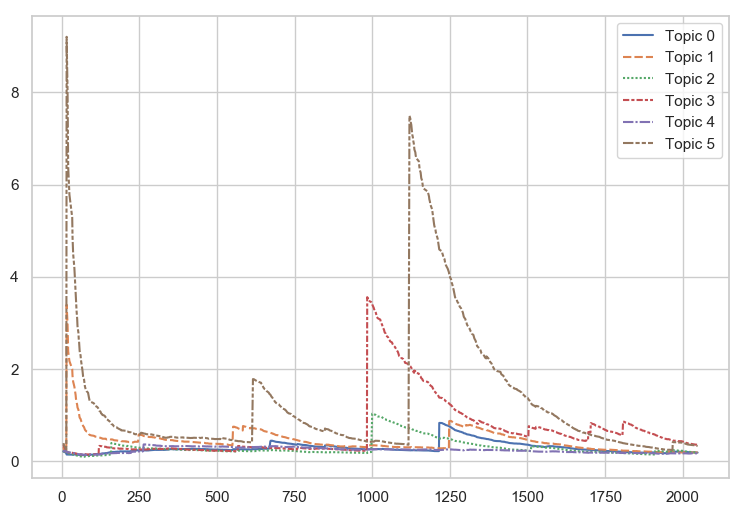}
  \caption{The last FC layer's weights' gradients}
  
\end{subfigure}
\caption{BN between the FC layers and at the $\beta$ matrix.}
\label{bn:yes-yes}
\end{figure}

To provide a deeper understanding of the training process of our model, we investigate the effect of BN on the \modelAbbr{}'s performance. In the model's architecture, we use BN at two components: between the fully connected (FC) layers and at the $\beta$ matrix. We carry out the experiment on the N20short dataset with 6 topics, where we alternately add and remove the BN at each component such that there are ultimately $4$ possible combinations. We plot the value of the Dirichlet parameters $\alpha$, the gradients of the topic-to-word distribution $\beta$ and the gradients of the weights of the last FC layer for each experiment.

Figure \ref{bn:no-no}, \ref{bn:no-yes}, \ref{bn:yes-no} and \ref{bn:yes-yes} depict the graphs of each combination of the 4 experiments. From the graph, we can observe that the removal of BN at the $\beta$ matrix leads to the divergence of the gradients between topics, where topics are divided into two groups: one group enjoys a high gradient while the other group obtains an increasingly smaller gradient. Therefore, the effect of BN at the $\beta$ matrix is to maintain a similar learning speed between topics in order to prevent the component collapsing phenomenon from happening. We can also witness the component collapsing phenomenon by observing the $\alpha$: its values diverge in Figure \ref{bn:no-no} and \ref{bn:yes-no} and remain similar in Figure \ref{bn:no-yes} and \ref{bn:yes-yes}.

On the other hand, the BN between the FC layers has the effect of allowing a larger gradient update at the weights of these layers. Figure \ref{bn:no-no} and \ref{bn:no-yes} show that without the BN at these layers, their gradient becomes extremely small, which means that the model fails to learn anything at all. By contrast, the gradients in Figure \ref{bn:yes-no} and \ref{bn:yes-yes} have a much larger magnitude throughout the training process.

We hypothesize that these phenomena occur due to the saturation of the softmax function during the training period. At the $\beta$ matrix, we use softmax to transform the original weights into a legitimate probability distribution. The training process may leads to some topics unfortunately having their softmax at the $\beta$ matrix saturated quickly, resulting in a diminishing gradient. The same event could happen with the Gumbel-Softmax distribution, where the saturation of the softmax means that the FC layers have very small gradients during backpropagation. Both of these catastrophic phenomena could be remedied by BN. Finally, BN is shown to smooth out the loss landscape \citep{NIPS2018_7515}, which allows the usage of a large learning rate and makes the model more robust against changes in hyperparameters.

\subsection{Discussion}
Overall, the \modelAbbr{} frequently returns better results than the original LDA model as well as the LFLDA on both the topic coherence evaluation and document clustering task. In terms of the significance of the word embedding, we found Word2Vec helps the model to reach better topic coherence scores than GloVe, while GloVe is evidently more beneficial to the document clustering task. This is actually quite logical, since the NPMI metric used in the topic coherence evaluation has a window-based nature similar to the Word2Vec. The NPMI employs a sliding window over tokens in an external corpus (Wikipedia) to calculate the cooccurrences between pairs of words, while the Word2Vec training method emphasizes capturing the relationships between a word and its neighbors, i.e., local co-occurrences. This is not to posit that the NPMI is a well-rounded metric for topic coherence, since it fails to take into account words that obviously come from the same topic but rarely occur near each other, such as names of competing brands, e.g., Lamborghini and Toyota. Conversely, the GloVe training method considers the cooccurrences within the entire document (global cooccurrences) and the indirect relationships between words, which is better in identifying words that belong to the same topic and grouping them together. In fact, table 3 in \cite{pennington_glove:_2014} suggests that GloVe is better than Word2Vec in word similarity tasks even when it is trained on a smaller corpus. This, in turn, leads to more words being included in one topic, which means more documents having the same label are assigned to the same cluster, reflected through a high agreement between the clustering result of the \modelAbbr{} and the ground truth label.

We consider how BN assists in the training process of our model. We conclude that the effect of BN is two-fold: first, it helps solve the component collapsing problem of the model caused by the rapid saturation of the softmax function; and second, it permits the employment of high learning rates which speeds up convergence.

Finally, the major advantage of our model is its speed. Our model can extract topics from a corpus using only a fraction of the time required by the LFLDA, and the time gap increases with a larger number of topics. Our model not only succeeds in securing a time advantage but also consumes less memory since it is trained using mini batches instead of loading the entire corpus into the memory, which makes it extremely useful for inducing topics of a large dataset of microtexts, e.g., posts on social media.

\section{Conclusion}
In this paper, we propose \modelAbbr{}, a model that can be combined with pretrained word vectors to induce latent topics from a dataset of microtexts.
We present a distribution \emph{q} that takes into account word embeddings to approximate the joint distribution of LDA using the variational inference method.
We express this distribution using a neural network for a faster convergence speed and a smaller memory footprint so that our model can operate on large datasets.
Experiments show the improved performance and runtime of our model in deriving latent topics from microtexts compared to other methods.
These advantages make our model suitable for real-world use cases, such as categorizing a large collection of comments on social networks.

\section*{Acknowledgments}
\subsection*{Conflict of interest}
We declare that we have no conflict of interest  for the work presented in this paper.

\bibliography{wileyNJD-APA}

\end{document}